\documentclass[runningheads]{llncs}

\usepackage{tikz}
\usepackage{comment}
\usepackage{amsmath,amssymb} 
\usepackage{color}
\usepackage{graphicx}
\usepackage{algorithm}
\usepackage{graphicx}
\usepackage{bm}
\usepackage{bbm}
\usepackage{gensymb}
\usepackage{color}
\usepackage{mathtools}
\usepackage{subfigure}

\usepackage{tikz}
\usepackage{comment}
\usepackage{amsmath,mathrsfs,amssymb} 
\usepackage{color}
\usepackage{algpseudocode}

\usepackage{multirow}
\usepackage{booktabs}
\usepackage[accsupp]{axessibility}  

\begin{document}
\pagestyle{headings}
\mainmatter
\def\ECCVSubNumber{2403}  

\title{CoupleFace: Relation Matters for Face Recognition Distillation} 

\titlerunning{CoupleFace}
%
\author{Jiaheng Liu\inst{1} \and
Haoyu Qin\inst{2*}\and
Yichao Wu\inst{2}\and Jinyang Guo\inst{3}\and \\Ding Liang\inst{2}\and Ke Xu\inst{1}}

\authorrunning{J. Liu et al.}
%
\institute{State Key Lab of Software Development Environment, Beihang University\and
SenseTime Group Limited\and SKLSDE, Institute of Artificial Intelligence, Beihang University
\email{liujiaheng@buaa.edu.cn},
\email{\{qinhaoyu1,wuyichao,liangding\}@sensetime.com}
}

\maketitle
\let\thefootnote\relax\footnotetext{*Corresponding author.}
 \begin{abstract}

{Knowledge distillation is an effective method to improve the performance of a lightweight neural network (i.e., student model)  by transferring the knowledge of a  well-performed neural network  (i.e.,  teacher model), which has been widely applied in many computer vision tasks, including face recognition (FR).
Nevertheless, the current FR distillation methods  usually utilize the Feature Consistency Distillation (FCD)  (e.g.,  $L_2$  distance)  on the learned embeddings extracted by the teacher and student models for each sample, which is not {able to fully} transfer the knowledge from the teacher to the student for FR. 
In this work, we observe that mutual relation knowledge between samples
is also important to improve the discriminative ability of the learned representation of the student model,
and propose an effective FR distillation method called CoupleFace by additionally introducing the Mutual Relation Distillation (MRD) into  existing distillation framework.
{Specifically,
in MRD,
we first propose to mine the informative mutual relations,
and then introduce the Relation-Aware Distillation (RAD) loss to transfer the mutual relation knowledge of the teacher model to the student model.}
Extensive experimental results on multiple benchmark datasets demonstrate the effectiveness of our proposed CoupleFace for FR. Moreover, based on our proposed CoupleFace, we have won the first place in the ICCV21 Masked Face Recognition Challenge (MS1M track).}
\end{abstract}

\section{Introduction}
Face recognition (FR) has been well investigated for decades.
Most of the progress
is credited to large-scale training datasets~\cite{Zhu_2021_CVPR,kemelmacher2016megaface}, resource-intensive networks with millions of parameters~\cite{he2016deep,simonyan2014very} and effective loss functions~\cite{deng2019arcface,wang2017normface}.
In practice, FR models are often deployed on  mobile and embedded devices, 
which are incompatible with the large neural networks (e.g., ResNet-101~\cite{he2016deep}).
\begin{figure}[!htp]
\begin{center}
\includegraphics[width=0.85\linewidth]{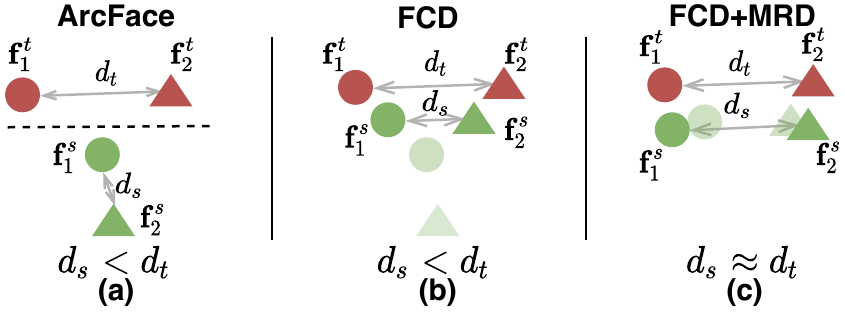}
\caption{The illustration of different methods for two samples from different classes. $\mathbf{f}^t_1$ and  $\mathbf{f}^t_2$ are extracted by teacher model, and $\mathbf{f}^s_1$ and  $\mathbf{f}^s_2$ are extracted by student model. $d_t$ and $d_s$ denote the distances  (i.e., $1-\mathrm{cos}(\mathbf{f}_1^t, \mathbf{f}_2^t)$ and $1-\mathrm{cos}(\mathbf{f}_1^s, \mathbf{f}_2^s)$) of the embeddings,
where $\mathrm{cos}(\cdot,\cdot)$ measures the cosine similarity of two features.
Model is better when distance is larger. (a). The teacher and student models are both trained by ArcFace~\cite{deng2019arcface}. (b). The teacher model is trained by ArcFace. The student model is trained by using FCD. (c). The teacher model is trained by ArcFace. The student model is trained by using both FCD and MRD in CoupleFace.}
\label{fig:intro}
\end{center}
\end{figure}
Besides, as shown in Fig.~\ref{fig:intro}(a), the capacities of the lightweight neural networks (e.g., MobileNetV2~\cite{sandler2018mobilenetv2}) cannot be fully exploited when they are only supervised by existing popular FR loss functions (e.g., ArcFace~\cite{deng2019arcface}).
Therefore,
how to develop lightweight and effective FR models  for real-world applications  has been investigated in recent years.
For example,
knowledge distillation~\cite{hinton2015distilling} is being actively discussed to produce lightweight and effective neural networks,
which transfers the knowledge from an effective teacher model to a lightweight student model.

 Most existing knowledge distillation works usually aim to guide the student to mimic the behavior of the teacher by introducing probability constraints (e.g., KL divergence~\cite{hinton2015distilling}) between the predictions of teacher and student models, which are not well-designed for FR.
In contrast,
as improving the discriminative ability of the feature embedding is the core problem for FR,
it is important to enable the student model to share the same embedding space with the teacher model for similarity comparison.
Thus, a simple and straightforward  FR distillation method  is to directly minimize the $L_2$ distance of the embeddings extracted by teacher and student models~\cite{wang2020exclusivity,shi2020proxylesskd},
which aims to align the embedding spaces between the teacher and student models.
We call this simple  method as Feature Consistency Distillation ({FCD}) as shown in Fig.~\ref{fig:intro}(b),
and FCD has been widely used in practice to improve the lightweight neural networks for FR.

In Fig.~\ref{fig:intro}(b),
 when FCD is used, feature embeddings extracted by student model get close to the corresponding feature embeddings extracted by teacher model.
 However, there are some cases that the distances of  embeddings (i.e., $d_s$) from different classes of 
 student model are still smaller than $d_t$ of teacher model.
 Similarly,
there are also some cases that  $d_s$ from the same class is  larger than $d_t$.
In practice, when student model is deployed, smaller $d_s$ for negative pair or larger $d_s$ for positive pair usually leads to false recognition,
which indicates that FCD is not sufficient to transfer the knowledge of teacher model to student model.
In Fig.~\ref{fig:intro}(c),
we define  the cosine similarity between samples as mutual relation and observe that $d_s \approx d_t$ when we additionally utilize the mutual relation information of  teacher model  to distill the corresponding mutual relation information of student model.
Thus,
we propose to introduce the Mutual Relation Distillation ({MRD}) into the existing FR distillation by reducing the gap between teacher and student models with respect to the mutual relation information.

Since FCD is able to align the embedding space of student model to teacher model well,
the student model can distinguish most image pairs easily,
which indicates that the differences of the most mutual relations between teacher and student models are relatively small.
In other words,
most mutual relations cannot provide valuable knowledge and affect the similarity distribution of all image pairs for FR. Therefore,
how to generate sufficient informative mutual relations efficiently in MRD is a challenging issue. 

Moreover,
recent metric learning works~\cite{harwood2017smart,wu2017sampling,suh2019stochastic} have  shown that hard negative samples are crucial for improving the discriminative ability of feature embedding.
But these works are not well-designed for mining mutual relations between samples for FR distillation.
To this end,
we propose to mine informative mutual relations in MRD.
Moreover, as the number of positive pairs is usually small, we focus on mining the mutual relations of negative pairs.
Overall, in our work,
we propose an effective FR distillation method referred to as CoupleFace, including FCD and MRD.
Specifically,
we first pre-train the teacher model on the large-scale training dataset.
Then,
in FCD,
we calculate the $L_2$ distance of the embeddings extracted by  the teacher and student models to generate the FCD loss for aligning the embedding spaces of the teacher and student models.
In MRD,
we first propose the Informative Mutual Relation Mining module to generate informative mutual relations efficiently in the training process,
where the informative prototype set generation and memory-updating strategies are introduced to improve the mining efficiency.
Then, we introduce the Relation-Aware Distillation (RAD) loss to exploit the mutual relation knowledge by using valid mutual relations and filtering out relations with subtle differences, 
which aims to better transfer the informative mutual relation knowledge from the teacher model to the student model.
The contributions  are summarized as follows:
\begin{itemize}
    \item In our work, we first investigate the importance of mutual relation knowledge for FR distillation, and propose an effective distillation framework called CoupleFace, which consists of Feature Consistency Distillation (FCD) and Mutual Relation Distillation (MRD).
    \item In MRD, we propose to obtain informative mutual relations efﬁciently in our Informative Mutual Relation Mining module, 
    where the informative prototype set generation and memory-updating strategies are used.
    Then, we introduce  the Relation-Aware Distillation (RAD) loss  to better transfer the mutual relation knowledge. 
    \item Extensive experiments on multiple benchmark datasets demonstrate the effectiveness and generalization ability of our proposed CoupleFace method.
\end{itemize}
\section{Related Work}
\noindent \textbf{Face Recognition.}
FR aims to maximize the inter-class discriminative ability and the intra-class compactness.
The success of FR can be summarized into the three factors:
effective loss functions~\cite{wang2017normface,zhang2017range,liu2017sphereface,sun2020circle,wang2018additive,deng2019arcface,deng2020sub,meng2021magface,Deng_2021_CVPR,liu2022anchorface,liu2021dam,li2020learning,oneface2022},
large-scale datasets~\cite{Zhu_2021_CVPR,An_2021_ICCV,Glint-Mini,yi2014learning,kemelmacher2016megaface},
 and
 powerful deep neural networks~\cite{taigman2014deepface,sun2014deep,simonyan2014very,Szegedy_2015_CVPR,liu2020block}.
The loss function design is the main-stream research direction for FR,
which  improves the generalization and discriminative abilities of the learned feature representation.
For example, Triplet loss~\cite{schroff2015facenet} is proposed to
enlarge the distances of negative pairs and reduce the distances of positive pairs.
Recently,
the angular constraint is applied into the cross-entropy loss function in 
many angular-based loss functions~\cite{liu2016large,liu2017sphereface}.
Besides, 
CosFace~\cite{wang2018cosface}
and ArcFace~\cite{deng2019arcface} further utilize a margin item for better discriminative capability of the feature representation.
Moreover,
 some mining-based loss functions (e.g., CurricularFace~\cite{huang2020curricularface} and MV-Arc-Softmax~\cite{wang2020mis}) take the difficulty degree of samples into consideration and achieve promising results.
The recent work VPL~\cite{Deng_2021_CVPR} additionally introduces the sample-to-sample comparisons to reduce the gap between the training and evaluation processes for FR.
In contrast, we propose to design an effective distillation loss function to improve the lightweight neural network.

\noindent\textbf{Knowledge Distillation.}
As a representative type of model compression and acceleration methods~\cite{guo2021jointpruning,guo2020multi,liu2022apsnet},
knowledge distillation aims to distill knowledge from a powerful teacher model into a lightweight student model~\cite{hinton2015distilling},
which has been applied in many computer vision tasks~\cite{yang2022cross,yang2021hierarchical,Peng_2019_ICCV,tian2019contrastive,park2019relational,huang2020improving,david2021margindistillation,fang2021seed,feng2020triplet,yang2022knowledge,liu2022C3,Huang_2022_CVPR}.
Many distillation methods have been proposed by utilizing different kinds of representation as knowledge for better performance. For example,
FitNet~\cite{romero2014fitnets} uses the middle-level hints from  hidden layers of the teacher model to guide the training process of the student model.
CRD~\cite{tian2019contrastive} utilizes a contrastive-based objective function for transferring knowledge between deep networks.
Some relation-based knowledge distillation methods (e.g., CCKD~\cite{Peng_2019_ICCV}, RKD~\cite{park2019relational}) utilize the relation knowledge to improve the student model.
Recently, knowledge distillation has also been applied to improve the performance of lightweight network (e.g., MobileNetV2~\cite{sandler2018mobilenetv2}) for FR.
For example,
EC-KD~\cite{wang2020exclusivity} proposes a position-aware exclusivity strategy to encourage diversity among different
filters of the same layer to alleviate the low capability of student models.
When compared with existing works,
CoupleFace is well-designed for  FR distillation by  considering to mine  informative mutual relations and transferring the relation knowledge  of the teacher model using Relation-Aware Distillation loss.

\begin{figure}[t]
\begin{center}
\includegraphics[width=0.6\linewidth]{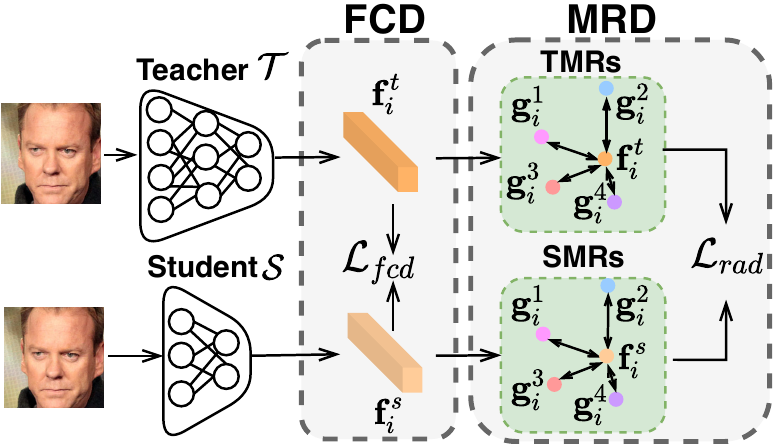}
\caption{The framework of CoupleFace for FR distillation. In FCD, we  use the $\mathcal{L}_{fcd}$ loss between $\mathbf{f}_i^s$ and $\mathbf{f}_i^t$ to align the embedding spaces of the teacher and student models. In MRD, we first mine the informative features  $\{\mathbf{g}_i^k\}_{k=1}^{K}$ (see Sec.~\ref{sec:imrm}),
where $K$ is the number of  features,
and calculate the teacher mutual relations (TMRs) and student mutual relations (SMRs) based  on $\mathbf{f}_i^s$, $\mathbf{f}_i^t$ and   $\{\mathbf{g}_i^k\}_{k=1}^{K}$.
Then, we minimize the $\mathcal{L}_{rad}$ loss to transfer the mutual relation knowledge of the teacher model to student model.
}
\label{fig:Overview}
\end{center}
\end{figure}

\section{Method}
In this section, we introduce the details of our CoupleFace in Fig.~\ref{fig:Overview},
which contains Feature Consistency Distillation (FCD) and Mutual Relation Distillation (MRD) for FR distillation. The overall pipeline is as follows.
First, we train the teacher model on a large-scale dataset.
Then,
in the distillation process of student model,
we extract the feature embeddings  based on the teacher and student models for each face image.
After that,
in FCD,
we compute the Feature Consistency Distillation (FCD) loss $\mathcal{L}_{fcd}$ based on $L_2$ distance of the feature embeddings.
Meanwhile,
in MRD,
we first build the informative mutual relations  using the feature embeddings of teacher and student models with the mined informative features as shown in Fig.~\ref{fig:imrm},
and then calculate the Relation-Aware Distillation (RAD) loss $\mathcal{L}_{rad}$ to transfer the mutual relation knowledge.
\subsection{Preliminary on Face Recognition Distillation}
In this section, we define some notations in CoupleFace, 
and discuss the necessity of FR distillation when compared with traditional knowledge distillation.

\noindent \textbf{Notations.}
We denote the teacher model as $\mathcal{T}$ and the student model as $\mathcal{S}$.
For each sample $x_i$,
the corresponding identity label is $y_i$,
and the corresponding features extracted by $\mathcal{T}$ and $\mathcal{S}$ are denoted as $\mathbf{f}_i^{t}$ and $\mathbf{f}_i^{s}$, respectively.

\noindent \textbf{Necessity of Face Recognition Distillation.}
For the traditional  knowledge distillation of image classification,
 the existing methods usually utilize the probability consistency~\cite{hinton2015distilling} (e.g., KL divergence) to align the prediction probabilities from  $\mathcal{S}$  with the prediction probabilities from $\mathcal{T}$.
However, the traditional knowledge distillation techniques are usually incompatible with FR.
In practice, for FR,
we can only obtain a pre-trained $\mathcal{T}$ but have no idea about how it was trained (e.g., the training datasets, loss functions).
Therefore, the probability consistency loss is not available when the number of identities of the training dataset for $\mathcal{T}$ is different from the current dataset for $\mathcal{S}$ or $\mathcal{T}$ is trained by other metric learning based loss functions (e.g., triplet loss~\cite{schroff2015facenet}).
Besides, FR models are trained to generate discriminative feature embeddings for  similarity comparison in the open-set setting rather than an effective classifier for the close-set classification. 
Thus, aligning the embedding spaces between  $\mathcal{S}$ and $\mathcal{T}$ is more important for FR distillation.
\subsection{Feature Consistency Distillation}
In Feature Consistency Distillation (FCD), to boost the performance of $\mathcal{S}$ for FR,
a simple and effective Feature Consistency Distillation (FCD) loss $\mathcal{L}_{fcd}$ is widely adopted in practice, which is defined as follows:
\begin{equation}
\small
    \mathcal{L}_{fcd} = \frac{1}{2N} \sum_{i=1}^{N}\bigg|\bigg|\frac{\mathbf{f}_i^t}{||\mathbf{f}_i^t||_2}-\frac{\mathbf{f}_i^s}{||\mathbf{f}_i^s||_2}\bigg|\bigg|^{2},
    \label{fcd}
\end{equation}
where $N$ is the number of face images for each mini-batch.



\subsection{Mutual Relation Distillation}
In this section,
we describe the  Mutual Relation Distillation (MRD) of CoupleFace in detail.
First, we discuss the necessity of MRD for FR distillation.
Then, we describe how to generate the  informative mutual relations by using our informative mutual relation mining strategy.
Finally,
we introduce the Relation-Aware Distillation (RAD) loss to transfer the mutual relation knowledge. 

\subsubsection{Necessity of Mutual Relation Distillation.}

First, we define a pair of embeddings as a couple.
Given a couple  $(\mathbf{f}_i^g, \mathbf{f}_j^g)$, $g \in \{s,t\}$ denotes $\mathcal{S}$ or $\mathcal{T}$,
the mutual relation $\mathrm{R}(\mathbf{f}_i^g, \mathbf{f}_j^g)$
of this couple is  defined as follows:
\begin{equation}
\small
    \mathrm{R}(\mathbf{f}_i^g, \mathbf{f}_j^g) = \mathrm{cos}(\mathbf{f}_i^g, \mathbf{f}_j^g),
    \label{couple}
\end{equation}
where $\mathrm{cos}(\cdot,\cdot)$ measures the cosine similarity between two features.
Given two couples $(\mathbf{f}_i^t, \mathbf{f}_i^s)$ and $(\mathbf{f}_j^t, \mathbf{f}_j^s)$,
when FCD loss is applied on $\mathcal{S}$,
the mutual relations $\mathrm{R}(\mathbf{f}_i^t, \mathbf{f}_i^s)$ and $\mathrm{R}(\mathbf{f}_j^t, \mathbf{f}_j^s)$ will be maximized.
However, in practice, when $\mathcal{S}$ is deployed, the mutual relation $\mathrm{R}(\mathbf{f}_i^s, \mathbf{f}_j^s)$ will be used to measure the similarity of this couple $(\mathbf{f}_i^s, \mathbf{f}_j^s)$ for FR.
Thus,
if we only use the FCD loss in FR distillation,
the optimization on the mutual relation $\mathrm{R}(\mathbf{f}_i^s, \mathbf{f}_j^s)$  is ignored,
which limits the further improvement of $\mathcal{S}$ for FR. 
Meanwhile, 
$\mathcal{T}$ with superior performance is able to provide an effective  mutual relation $\mathrm{R}(\mathbf{f}_i^t, \mathbf{f}_j^t)$ as the ground-truth to distill the mutual relation  $\mathrm{R}(\mathbf{f}_i^s, \mathbf{f}_j^s)$  from $\mathcal{S}$.
Therefore, we introduce the MRD into the existing FR distillation framework.

However, direct optimization on mutual relation $\mathrm{R}(\mathbf{f}_i^s, \mathbf{f}_j^s)$ may not be a good choice in practice. Due to the batch size limitation and randomly sampling strategy in the training process,
the mutual relations for these couples $\{(\mathbf{f}_i^s, \mathbf{f}_j^s)\}_{j=1, j\neq i}^{N}$ constructed across the mini-batch cannot  support an effective and efficient mutual relation distillation. To this end, in CoupleFace, we propose to optimize $\mathrm{R}(\mathbf{f}_i^s, \mathbf{f}_j^t)$  instead of $\mathrm{R}(\mathbf{f}_i^s, \mathbf{f}_j^s)$ for following reasons.
First, the quantity of mutual relations for these couples $\{(\mathbf{f}_i^s, \mathbf{f}_j^t)\}_{j=1, j\neq i}^{L}$ can be very large,
where $L$ is the number of samples of the dataset and $\mathbf{f}_j^t$ can be pre-calculated using teacher model.
Second, the quality of the mutual relation for couple $(\mathbf{f}_i^s, \mathbf{f}_j^t)$ can be guaranteed as it can be mined from sufficient couples.
Third, $\mathbf{f}_j^s$ is almost the same as  $\mathbf{f}_j^t$ from the perspective of $\mathbf{f}_i^s$ when the FCD loss between $\mathbf{f}_j^s$ and $\mathbf{f}_j^t$  is fast converged, which represents that the mutual relation $\mathrm{R}(\mathbf{f}_i^s, \mathbf{f}_j^t)$ of couple $(\mathbf{f}_i^s, \mathbf{f}_j^t)$ is an ideal approximation of $\mathrm{R}(\mathbf{f}_i^s, \mathbf{f}_j^s)$ of couple $(\mathbf{f}_i^s, \mathbf{f}_j^s)$. Therefore, we propose to use the mutual relation $\mathrm{R}(\mathbf{f}_i^t, \mathbf{f}_j^t)$  to distill $\mathrm{R}(\mathbf{f}_i^s, \mathbf{f}_j^t)$.

Note that we call $\mathrm{R}(\mathbf{f}_i^t, \mathbf{f}_j^t)$ and $\mathrm{R}(\mathbf{f}_i^s, \mathbf{f}_j^t)$ as teacher mutual relation (\textbf{TMR}) and student mutual relation (\textbf{SMR}), respectively.
To sum up,
in MRD,
we propose to utilize TMRs to distill the SMRs in the training process of FR.
\subsubsection{Informative Mutual Relation Mining.}
\label{sec:imrm}
\begin{figure}[t]
\begin{center}
\includegraphics[width=0.9\linewidth]{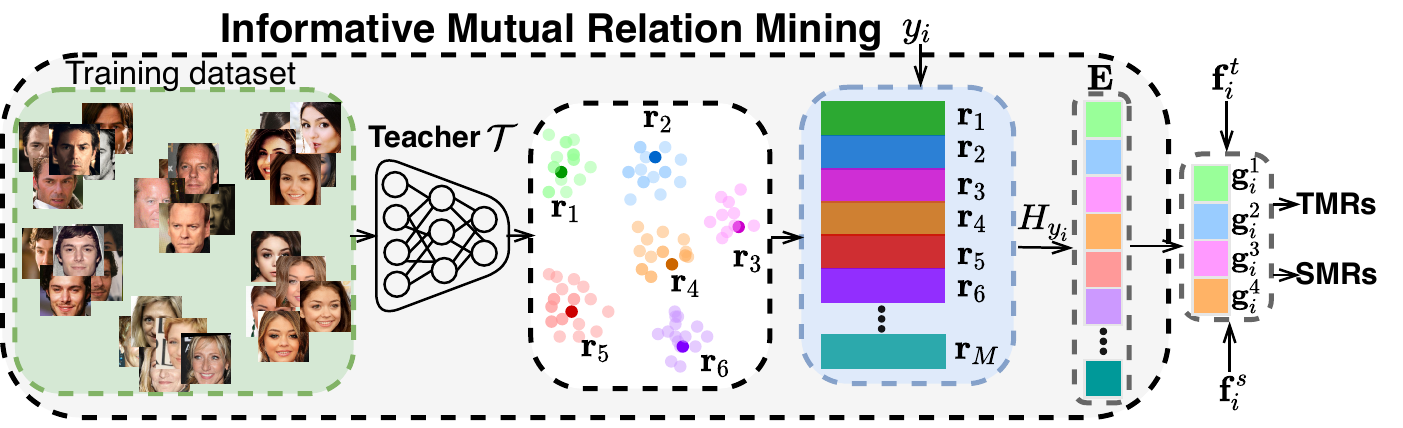}
\caption{We first pre-calculate identity prototypes $\{\mathbf{r}_m\}_{m=1}^M$ and generate the informative prototype set $H_{y_i}$ for identity $y_i$ using our informative prototype set generation strategy, where $M$ is the number of identities across the training dataset.
Then, we maintain a feature bank $\mathbf{E}\in \mathbb{R}^{M\times d}$ to store the feature embeddings of $\mathcal{T}$, which is updated in each iteration using our memory-updating strategy.
Finally, we obtain $K$ informative features $\{\mathbf{g}_i^k\}_{k=1}^K$ based on $H_{y_i}$ and $\mathbf{E}$,
and construct the SMRs and the TMRs based on $\mathbf{f}_i^s$, $\mathbf{f}_i^t$ and $\{\mathbf{g}_i^k\}_{k=1}^K$. Here, we set $K=4$ for better illustration.
}
\label{fig:imrm}
\end{center}
\end{figure}
In this section,
we describe how to generate informative mutual relations efficiently  in Fig.~\ref{fig:imrm} of CoupleFace.

Intuitively, for $\mathbf{f}_i^s$ and $\mathbf{f}_i^t$,
a straightforward way is to construct the couples across all training samples 
and generate SMRs 
and TMRs.
However, the computation cost is very large in this way,
which is not applicable in practice.
An alternative way is to generate  SMRs 
and 
TMRs 
across the mini-batch.
However, the batch size is relatively small and most mutual relations cannot provide valuable knowledge to improve $\mathcal{S}$,
as it is easy to distinguish most image pairs and only the hard image pairs will greatly affect the performance of FR model.
Meanwhile,
recent metric learning works~\cite{harwood2017smart,wu2017sampling,suh2019stochastic} show that hard negative samples are crucial for improving the discriminative ability of the embeddings.
Therefore, in MRD,
we propose to mine informative mutual relations among negative pairs to reduce the computation cost and improve $\mathcal{S}$ as shown in Fig.~\ref{fig:imrm}.
Specifically, we use an informative prototype set generation strategy to find a set of most similar identities called informative prototype set $H_{y_i}$ for identity $y_i$. Then, we utilize a memory-updating strategy to build the informative mutual relations based on $\mathbf{f}_i^s$, $\mathbf{f}_i^t$ and the features belonging to $H_{y_i}$ efficiently.

\noindent\textbf{Informative prototype set generation.}
First, we extract the features of the training data by using a well-performed trained model.
In our work, we directly use $\mathcal{T}$.
The generated features can be denoted as $\{\mathbf{f}_i^{t}\}_{i=1}^L$ and the corresponding identity label is $y_i$ for $\mathbf{f}_i^{t}$.
For the training dataset,
we denote the number of samples  as $L$,
the number of identities  as $M$,
and the identity label set as $\{m\}_{m=1}^M$.
For each identity $m$,
we calculate the identity  prototype $\mathbf{r}_{m}$ as follows:
\begin{equation}
\small
    \mathbf{r}_m = \frac{1}{l_m}\sum_{i=1,\\y_i = m}^{L}{\frac{\mathbf{f}_i^t}{||\mathbf{f}_i^t||_2}},
    \label{center}
\end{equation}
where $l_m$ is the number of samples in the training dataset for identity $m$.
Then, all identity prototypes of the dataset can be denoted as $\{\mathbf{r}_m\}_{m=1}^M$.
After that,
to find the informative prototype set  $H_{m}$ for $\mathbf{r}_m$,
 we calculate the cosine similarity between $\mathbf{r}_m$ and $\mathbf{r}_n$, where $n\in \{m\}_{m=1}^{M}$ and $n\neq m$.
 Afterwards, we select the top $K$ (e.g., $K=100$) identities with the largest similarities to construct $H_{m}$ for identity $m$,
 where  $H_{m}$ contains $K$ identity labels.
 Finally, the informative prototype set  for identity $y_i$ is denoted as $H_{y_i}$.
 
 \noindent\textbf{Memory-updating.}
 As the number of samples belonging to $H_{y_i}$ is also relatively large,
 we further propose a memory-updating strategy to reduce the computation cost while making full use of all samples  belonging to $H_{y_i}$ inspired by MoCo~\cite{he2020momentum}.
 Specifically,
 we maintain a feature bank $\mathbf{E}\in \mathbb{R}^{M\times d}$ to store feature embeddings extracted by $\mathcal{T}$, where only one embedding is preserved for each identity and $d$ is the dimension (e.g., 512) of the embedding.
At the beginning of the training process for $\mathcal{S}$, we initialize the feature bank $\mathbf{E}$ by randomly selecting one feature embedding generated by $\mathcal{T}$ for each identity.
Then, in each iteration,
 we first obtain the  feature embeddings $\{\mathbf{f}^t_i\}_{i=1}^{N}$ extracted by $\mathcal{T}$,
 where $N$ is the size of mini-batch,
 and we update the feature bank $\mathbf{E}$ by setting $\mathbf{E}[y_i]=\mathbf{f}_i^t$,
 where $[\cdot]$ means to obtain  features from $\mathbf{E}$ based on identity $y_i$.

Based on the informative prototype set $H_{y_i}$  for $y_i$, we can obtain $K$ informative negative features $\mathbf{G}_i =  \mathbf{E}[H_{y_i}]$,
where $\mathbf{G}_i \in \mathbbm{R}^{K \times d}$.
Meanwhile, we denote  each feature in $\mathbf{G}_i$ as $\mathbf{g}_i^k$,
where $k\in\{1,...,K\}$.
Finally,
a set of couples $\{(\mathbf{f}_i^s, \mathbf{g}_i^k)\}_{k=1}^K$ is constructed for $\mathbf{f}_i^s$ and  we can calculate the informative SMRs using these couples.
Similarly,
we can also generate the informative TMRs based on a set of couples $\{(\mathbf{f}_i^t, \mathbf{g}_i^k)\}_{k=1}^K$ as the ground-truth of these SMRs.
\subsubsection{Relation-Aware Distillation Loss.}
Based on the mined TMRs and SMRs,
the Relation-Aware Distillation (\textbf{RAD}) loss can be easily defined as follows:
\begin{equation}
\small
   \mathcal{L}_{rad} = \frac{1}{NK} \sum_{i=1}^N\sum_{k=1}^K{|\mathrm{cos}(\mathbf{f}_i^s, \mathbf{g}_{i}^k)-\mathrm{cos}(\mathbf{f}_i^t, \mathbf{g}_{i}^k)|}.
    \label{L_R1}
\end{equation}

However, the teacher model is not always better than the
student model for each case in the training dataset.
As illustrated in Fig.~\ref{vis} of Sec.~\ref{ana},
 we observe that $\mathrm{cos}(\mathbf{f}_i^t, \mathbf{g}_{i}^k) > \mathrm{cos}(\mathbf{f}_i^s, \mathbf{g}_{i}^k)$ does exist between the mined TMRs and SMRs. 
Therefore,
if we directly use the Eq.~(\ref{L_R1}) to transfer the mutual relation knowledge of $\mathcal{T}$,
$\mathcal{S}$ will be misled in some cases, which may degrade the performance of $\mathcal{S}$. 

To this end, we propose only to  use the valid mutual relations when  $\mathrm{cos}(\mathbf{f}_i^t, \mathbf{g}_{i}^k) < \mathrm{cos}(\mathbf{f}_i^s, \mathbf{g}_{i}^k)$,
and we reformulate the RAD loss of Eq.~(\ref{L_R1})  as follows:
\begin{equation}
\small
   \mathcal{L}_{rad} = \frac{1}{N^{'}} \sum_{i=1}^N\sum_{k=1}^{K}{\mathrm{max}(\mathrm{cos}(\mathbf{f}_i^s, \mathbf{g}_{i}^k)-\mathrm{cos}(\mathbf{f}_i^t, \mathbf{g}_{i}^k), 0)},
    \label{L_R2}
\end{equation}
where $N^{'}$ is the number of valid mutual relations across the mini-batch.
Thus,
RAD loss of Eq.~(\ref{L_R2}) will only affect the gradient when  $\mathrm{cos}(\mathbf{f}_i^t, \mathbf{g}_{i}^k) < \mathrm{cos}(\mathbf{f}_i^s, \mathbf{g}_{i}^k)$ and transfer the accurate mutual relation knowledge from $\mathcal{T}$ to $\mathcal{S}$ in MRD.
We mine informative mutual relations for each identity.
But there exists some identities, which can be easily distinguished from other identities,
which indicates that the differences between the mined SMRs and TMRs for these identities are still subtle.
Inspired by the hinge loss~\cite{gentile1998linear}, we further propose a more effective  variant of our RAD loss by introducing a margin $q$ as follows:
\begin{equation}
\small
  \mathcal{L}_{rad} = \frac{1}{N^{'}} \sum_{i=1}^N\sum_{k=1}^{K}{\mathrm{max}(\mathrm{cos}(\mathbf{f}_i^s, \mathbf{g}_{i}^k)-\mathrm{cos}(\mathbf{f}_i^t, \mathbf{g}_{i}^k)-q, 0)}.
    \label{L_R3}
\end{equation}
Intuitively,
the RAD loss in Eq.~(\ref{L_R3}) further filters out the mutual relations with subtle differences between the SMRs and TMRs and pays attention to these SMRs, which are far away from their corresponding TMRs.


\subsection{Loss Function of CoupleFace}
The overall loss function of CoupleFace is defined as follows:
\begin{equation}
\small
    \mathcal{L} = \mathcal{L}_{fcd} + \alpha \cdot \mathcal{L}_{rad} +\beta \cdot \mathcal{L}_{ce},
    \label{overall}
\end{equation}
where $\alpha$ and $\beta$ are the weights of  RAD loss $\mathcal{L}_{rad}$ and recognition loss $\mathcal{L}_{ce}$ (e.g., ArcFace~\cite{deng2019arcface}), respectively.
We also provide an algorithm in Alg.~\ref{alg:pseudocode}.

\begin{algorithm}[t]
\caption{CoupleFace}
\label{alg:pseudocode}
\begin{algorithmic}[1]
\Require
   Pre-trained teacher model $\mathcal{T}$; 
   Randomly initialized student model $\mathcal{S}$; 
   Current batch with $N$  images;
   The dimension of feature representation $d$;
   The number of identities $M$;
   The training dataset with $L$ images;
   The feature bank $\mathbf{E}\in \mathbb{R}^{M\times d}$;
\State Extract all features $\{\mathbf{f}^t_i\}_{i=1}^{L}$ of the dataset using $\mathcal{T}$;
\State Generate all  identity prototypes $\{\mathbf{r}_m\}_{m=1}^{M}$ based on $\{\mathbf{f}^t_i\}_{i=1}^{L}$ according to Eq.~(\ref{center});
\State Based on $\{\mathbf{r}_m\}_{m=1}^{M}$, calculate informative prototype set $\{H_{y_i}\}_{i=1}^L$ for each sample according to the label $y_i$;
\State Initialize feature bank $\mathbf{E}$ using $\{\mathbf{f}^t_i\}_{i=1}^{L}$;
\For {each iteration in the training process}
\State Get features $\{\mathbf{f}^t_i\}_{i=1}^{N}$ extracted by  $\mathcal{T}$ from $\{\mathbf{f}^t_i\}_{i=1}^{L}$;
\State Get features $\{\mathbf{f}^s_i\}_{i=1}^{N}$ extracted by $\mathcal{S}$;
\State Calculate $\mathcal{L}_{fcd}$ of $\{\mathbf{f}^s_i\}_{i=1}^{N}$ and $\{\mathbf{f}^t_i\}_{i=1}^{N}$ by Eq.~(\ref{fcd});
\State Update feature bank $\mathbf{E}$ using $\{\mathbf{f}^t_i\}_{i=1}^{N}$;
\For{each feature $\mathbf{f}_i^s$ in $\{\mathbf{f}_i^s\}_{i=1}^{N}$}
\State Get $K$ informative negative features $\{\mathbf{g}_i^k\}_{k=1}^{K}$  from   $\mathbf{E}$  using $H_{y_i}$;
\State  Build TMRs and SMRs by $\{(\mathbf{f}_i^s, \mathbf{g}_i^k)\}_{k=1}^K$ and $\{(\mathbf{f}_i^t, \mathbf{g}_i^k)\}_{k=1}^K$, respectively;
\EndFor
\State Calculate $\mathcal{L}_{rad}$ using TMRs and SMRs by Eq.~(\ref{L_R3});
\State Update the parameters of $\mathcal{S}$  by minimizing the loss  function   \Statex \quad \ \ $\mathcal{L} = \mathcal{L}_{fcd} + \alpha \cdot \mathcal{L}_{rad} + \beta \cdot \mathcal{L}_{ce} $;
\EndFor
\Ensure
The optimized student model $\mathcal{S}$.
\end{algorithmic}
\end{algorithm}
\section{Experiments}

\noindent\textbf{Datasets.}
For training,  the mini version of Glint360K~\cite{An_2021_ICCV} named as Glint-Mini \cite{Glint-Mini} is used,
where Glint-Mini \cite{Glint-Mini} contains 5.2M images of 91k identities.
For testing,
we use four datasets (i.e.,
  IJB-B~\cite{whitelam2017iarpa}, IJB-C~\cite{maze2018iarpa},
  and MegaFace~\cite{kemelmacher2016megaface}).

\noindent\textbf{Experimental setting.}
For the pre-processing of the training data,
we follow the recent works~\cite{deng2019arcface,kim2020broadface,deng2020sub} to generate the normalized face crops ($112\times112$).
For teacher models,
we use the widely used large neural networks (e.g.,
ResNet-34, ResNet-50 and ResNet-100~\cite{he2016deep}).
For student models,
we use MobileNetV2~\cite{sandler2018mobilenetv2} and ResNet-18~\cite{he2016deep}.
For all models,
the feature dimension  is 512.
For the training process of all models based on ArcFace loss,
the initial learning rate is 0.1 and divided by 10 at the 100k, 160k, 180k iterations.
The batch size and the total iteration are set as 512 and 200k, respectively.
For the distillation process,
the initial learning rate is 0.1 and divided by 10 at the 45k, 70k, 90k iterations.
The batch size and the total iteration are set as 512 and 100k, respectively.
In the informative mutual relation mining stage,
we set the number of most similar identities (i.e., $K$) as 100.
In Eq.~(\ref{L_R3}), we set the margin (i.e., $q$) as 0.03. The loss weight $\alpha$ is set as 1,
where $\beta$ is set as 0 in the first 100k iterations,
and is set as 0.01 in CoupleFace+ of Table~\ref{ijb}.
In the following experiments of different distillation methods, by default, we use the ResNet-50 ({R-50}), MobileNetV2 ({MBNet}) as $\mathcal{T}$ and $\mathcal{S}$, respectively.
\subsection{Results on the IJB-B and IJB-C datasets}
As shown in Table~\ref{ijb},
the first two rows represent the performance of models trained by using the ArcFace loss function~\cite{deng2019arcface}.
We compare our method with classical KD~\cite{hinton2015distilling}, FCD, 
CCKD~\cite{Peng_2019_ICCV}, SP~\cite{Tung_2019_ICCV}, RKD~\cite{park2019relational}, 
EC-KD~\cite{wang2020exclusivity}.
For FCD,
we only use the FCD loss of Eq.~(\ref{fcd}) to align the embedding space of the student and teacher models,
which is a very strong baseline to improve the performance of student model for FR.
For these methods (i.e., CCKD~\cite{Peng_2019_ICCV}, SP~\cite{Tung_2019_ICCV} and RKD~\cite{park2019relational}),
we combine these methods with FCD loss instead of the classical KD loss to achieve better performance.
For EC-KD proposed for FR, we reimplement this method.
In Table~\ref{ijb},
FCD is much better than classical KD,
which indicates the importance of aligning embedding space for FR when compared with classical KD.
Moreover,
we observe that CoupleFace achieves significant performance improvements when compared with existing methods,
which demonstrates the effectiveness of CoupleFace.
For the CoupleFace+,
we first pretrain student by CoupleFace,
and then train student by CoupleFace with ArcFace by setting $\beta$ in Eq.(~\ref{overall}) as 0.01 for another 100k iterations,
better results are obtained.
        \begin{table}[t]
            \centering
    \caption{Results (TAR@FAR) on IJB-B and IJB-C of different methods.}
         \label{ijb}

          \begin{tabular}{c|c|cc|cc}
\toprule
            \multirow{2}{*}{Models}&
               \multirow{2}{*}{Method}&
               \multicolumn{2}{c|}{IJB-B}&
               \multicolumn{2}{c}{IJB-C }\\
              \cline{3-6}
               && 1e-4 & 1e-5 & 1e-4 &1e-5\\
               \hline
               R-50~\cite{he2016deep}&ArcFace~\cite{deng2019arcface} &93.89&89.61&95.75&93.44\\
               MBNet~\cite{sandler2018mobilenetv2}&ArcFace~\cite{deng2019arcface}&85.97&75.81&88.95&82.64\\\hline
                \multirow{8}{*}{MBNet~\cite{sandler2018mobilenetv2}}
                &KD~\cite{hinton2015distilling}&86.12&75.99&89.03&82.69\\
                &FCD&90.34&81.92&92.68&87.74\\
                &CCKD~\cite{Peng_2019_ICCV}&90.72&83.34&93.17&89.11\\
                &RKD~\cite{park2019relational}&90.32&82.45&92.33&88.12\\
                &SP~\cite{Tung_2019_ICCV}&90.52&82.88&92.71&88.52\\
                &EC-KD~\cite{wang2020exclusivity}&90.59&83.54&92.85&88.32\\
                &CoupleFace&{91.18}&{84.63}&{93.18}&{89.57}\\
                &CoupleFace+&\textbf{91.48}&\textbf{85.12}&\textbf{93.37}&\textbf{89.85}\\
               \bottomrule
 \end{tabular}
        \end{table}

\subsection{Results on the MegaFace dataset}
In Table~\ref{tab2},
we also provide the results of CoupleFace on MegaFace~\cite{wolf2011face},
and we observe that CoupleFace is better than other methods.
For example,
when compared with the FCD baseline,
our method  improves the rank-1 accuracy by 0.62\% on MegaFace  under the distractor size as $10^6$.

\begin{table}[!htp]
    \centering
            \caption{Rank-1  accuracy with different distractors on MegaFace.}
    \footnotesize
    \begin{tabular}{c|c|ccc}
        \toprule
                    \multirow{2}{*}{Models}&
              \multirow{2}{*}{Method}&
              \multicolumn{3}{c}{Distractors}\\
                             \cline{3-5}
                            & &\scalebox{0.9}{$10^4$}&\scalebox{0.9}{$10^5$}&\scalebox{0.9}{$10^6$}\\
        \hline
              R-50~\cite{he2016deep} &ArcFace~\cite{deng2019arcface}&99.40&98.98&98.33\\
              MBNet~\cite{sandler2018mobilenetv2}&ArcFace~\cite{deng2019arcface}&94.56&90.25&84.64\\\hline
                 \multirow{7}{*}{MBNet~\cite{sandler2018mobilenetv2}}
                &KD~\cite{hinton2015distilling}&94.46&90.25&84.65\\
                &FCD&97.81&96.39&93.65\\
                &CCKD~\cite{Peng_2019_ICCV}&98.07&96.43&93.90\\
                &RKD~\cite{park2019relational}&98.06&96.41&93.84\\
                &SP~\cite{Tung_2019_ICCV}&98.01&96.58&93.95\\
                &EC-KD~\cite{wang2020exclusivity}&98.00&96.41&93.85\\
                &CoupleFace&\textbf{98.09}&\textbf{96.74}&\textbf{94.27}\\     
            \bottomrule
    \end{tabular}

    \label{tab2}
\end{table}
\subsection{Ablation Study}
\noindent \textbf{The effect of different variants of RAD loss.}
In MRD,
we propose three variants of RAD loss (i.e., Eq.~(\ref{L_R1}), Eq.~(\ref{L_R2}) and Eq.~(\ref{L_R3})).
To analyze the effect of different variants,
we also perform additional experiments based on Eq.~(\ref{L_R1}) and Eq.~(\ref{L_R2}) and report the results of MBNet on IJB-B and IJB-C.
In Table~\ref{variants},
for CoupleFace-A,
we replace the RAD loss of Eq.~(\ref{L_R3}) with Eq.~(\ref{L_R1}),
which means that we transfer the mutual relations by distilling all TMRs to the corresponding SMRs without any selection process.
For CoupleFace-B,
we replace the RAD loss of Eq.~(\ref{L_R3}) with Eq.~(\ref{L_R2}) without using the margin item.
In Table~\ref{variants},
we observe that CoupleFace-B outperforms CoupleFace-A a lot,
which shows the effectiveness of only using the mutual relations when $\mathrm{cos}(\mathbf{f}_i^t, \mathbf{g}_{i}^k) < \mathrm{cos}(\mathbf{f}_i^s, \mathbf{g}_{i}^k)$.
Moreover, CoupleFace also outperforms CoupleFace-B,
so it is necessary to emphasize the distillation on these SMRs,
which are far from their corresponding TMRs.
\begin{table}[t]
        \centering
  \caption{Results on IJB-B and IJB-C of different methods.}
   \label{variants}
          \begin{tabular}{c|cc|cc}
\toprule
               \multirow{2}{*}{Methods}& \multicolumn{2}{c|}{IJB-B}&\multicolumn{2}{c}{IJB-C}\\
               \cline{2-5}   
                &1e-4 &1e-5&1e-4&1e-5\\\hline
                CoupleFace &91.18&84.63&93.18&89.57\\
                CoupleFace-A &90.73&83.68&92.75&88.35\\
                CoupleFace-B &90.88&84.23&93.02&88.99\\
                        CoupleFace-C &90.65&83.18&92.52&88.89\\
                CoupleFace-D &90.85&83.73&92.86&89.04\\
                CoupleFace-E &90.78&83.32&92.78&88.79\\
               \bottomrule
 \end{tabular}
        \end{table}

\noindent \textbf{The effect of Informative Mutual Relation Mining.}
To demonstrate the effect of mining the informative mutual relations,
we further propose three alternative variants of CoupleFace (i.e., CoupleFace-C, CoupleFace-D, CoupleFace-E).
Specifically,
for CoupleFace-C,
we propose to directly optimize mutual relation $\mathrm{R}(\mathbf{f}_i^s, \mathbf{f}_j^s)$ without the  process of mining mutual relations,
where only mutual relations from limited couples $\{(\mathbf{f}_i^s, \mathbf{f}_j^s)\}_{j=1, j\neq i}^{N}$ across the mini-batch are constructed.
For CoupleFace-D,
we randomly select $K=100$ identities to construct the $H_{y_i}$ for each identity $y_i$,
while for CoupleFace-E, we propose to build the TMRs and SMRs across the mini-batch without mining process, and compute the RAD loss from them.
As shown in Table~\ref{variants},
we observe that CoupleFace is much better than three alternative variants,
which demonstrates that  it is beneficial to mine the informative mutual relations in CoupleFace.

\noindent
\textbf{The effect of informative  prototype set generation when using different models.}
In our work, we directly use $\mathcal{T}$ (i.e., R-50) to generate  the informative prototype set $H_{y_i}$ for each identity $y_i$.
Here, we propose to use other pre-trained models (i.e., \textbf{MBNet} and ResNet-100 (\textbf{R-100})) trained by ArcFace loss to generate $H_{y_i}$ for $y_i$,
and we call these alternative methods as CoupleFace-MBN and CoupleFace-RN100, respectively.
In Table~\ref{net},
we  report the results of MBNet  on IJB-B and IJB-C after using CoupleFace, CoupleFace-MBN and CoupleFace-RN100.
We observe that CoupleFace achieves comparable performance with CoupleFace-RN100,
and higher performance than CoupleFace-MBN.
For this phenomenon,
we assume that when using a more effective model,
we will generate more discriminative identity prototypes $\{\mathbf{r}_m\}_{m=1}^M$,
which leads to generating a more accurate informative  prototype set $H_{y_i}$.
Thus, it is beneficial to use effective models for obtaining the informative prototype set.

        \begin{table}[t]
            \centering
         \caption{Results on IJB-B and IJB-C of different methods.}
          \begin{tabular}{c|cc|cc}
\toprule
               \multirow{2}{*}{Methods}& \multicolumn{2}{c|}{IJB-B}&\multicolumn{2}{c}{IJB-C}\\
               \cline{2-5}   
                &1e-4 &1e-5&1e-4&1e-5\\\hline
                CoupleFace &91.18&84.63&93.18&89.57\\
                CoupleFace-MBN &91.06&83.91&92.95&88.94\\
                CoupleFace-RN100 &91.15&84.65&93.16&89.58\\
               \bottomrule
 \end{tabular}
    \label{net}
\end{table}
\subsection{Further Analysis}
\label{ana}
\begin{figure}[t]
    \centering
    \subfigure{
        \includegraphics[width=0.25\linewidth]{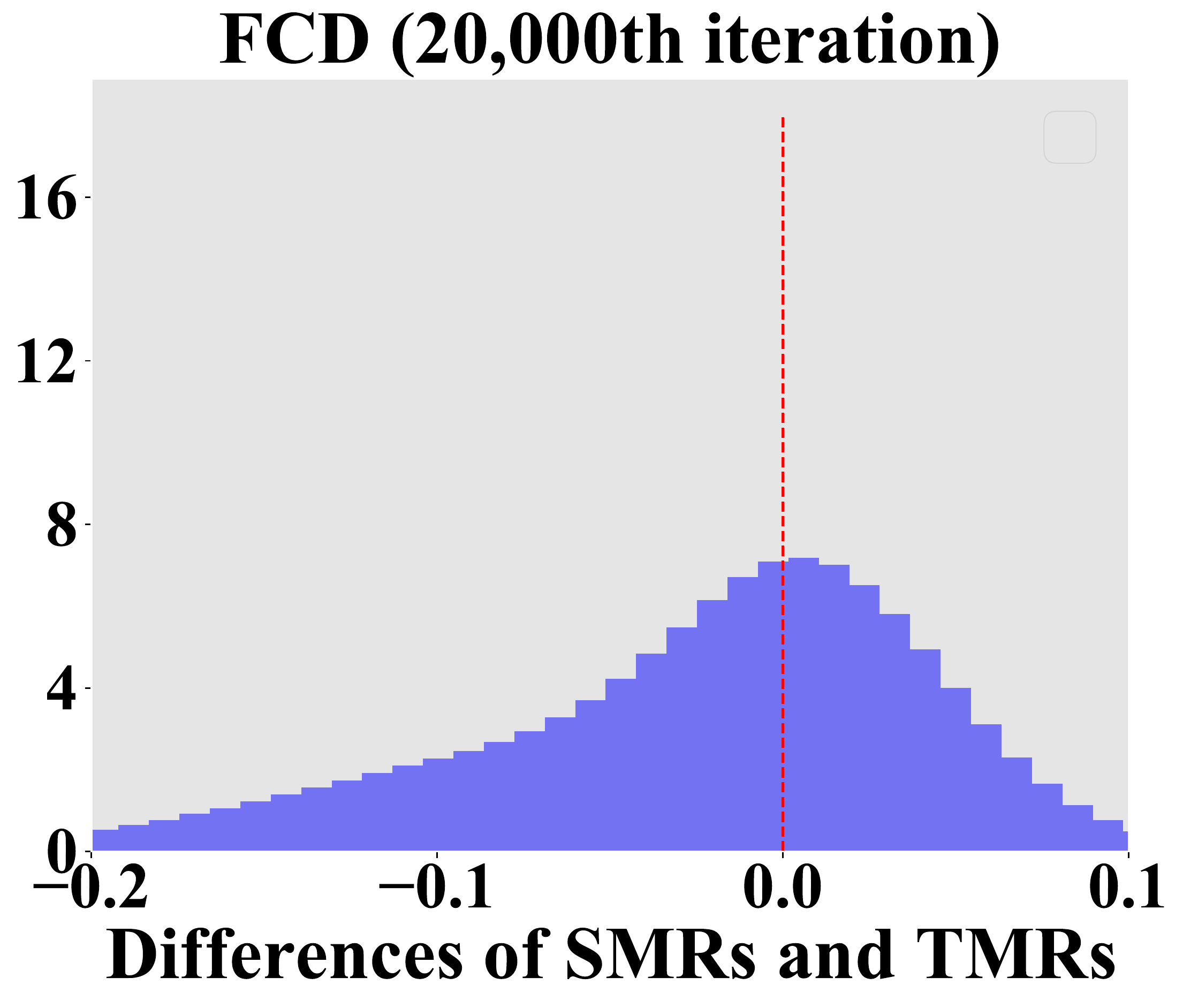}
        \label{label_for_cross_ref_1}
    }
    \subfigure{
	\includegraphics[width=0.25\linewidth]{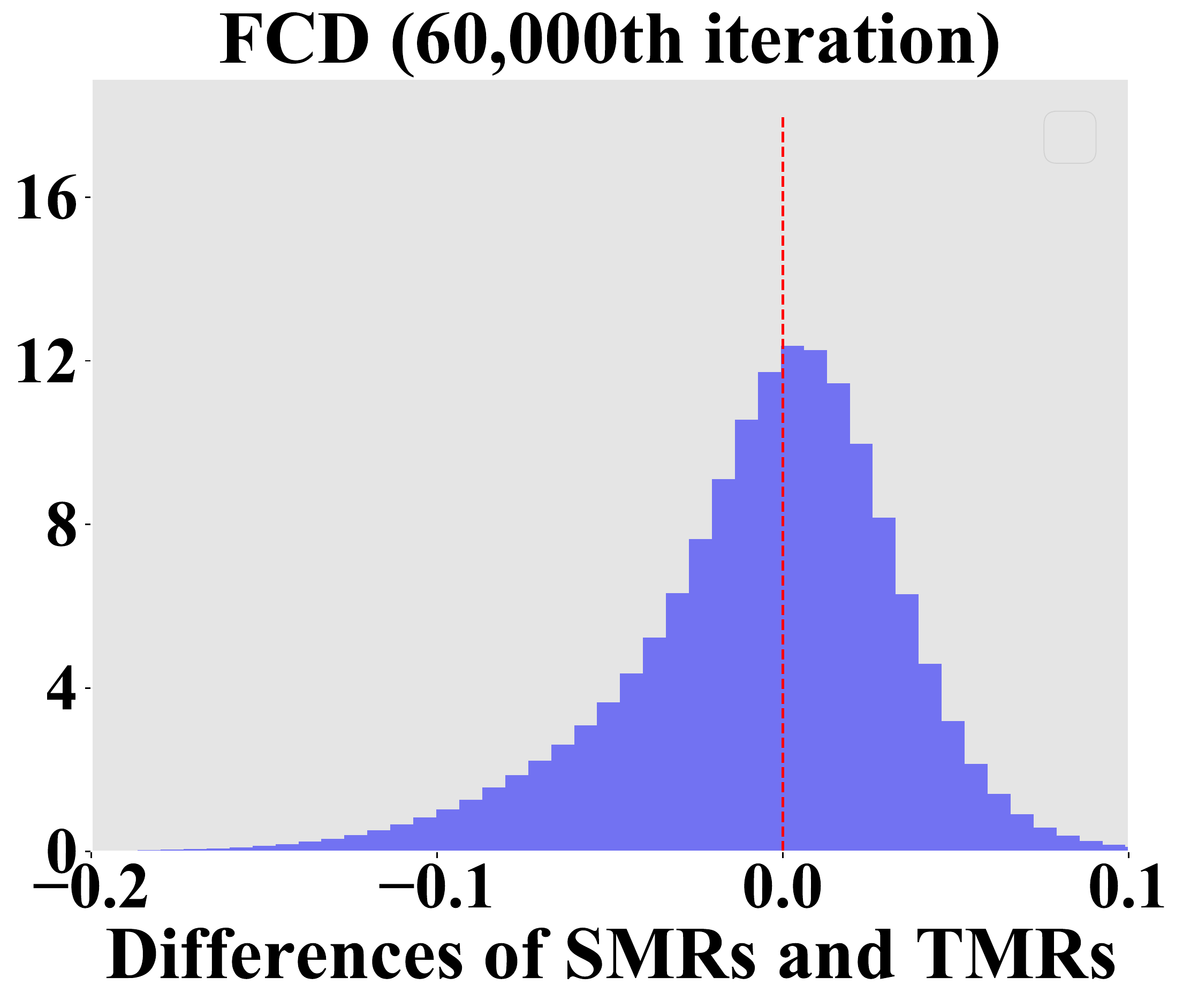}
        \label{label_for_cross_ref_2}
    }
    \subfigure{
    	\includegraphics[width=0.25\linewidth]{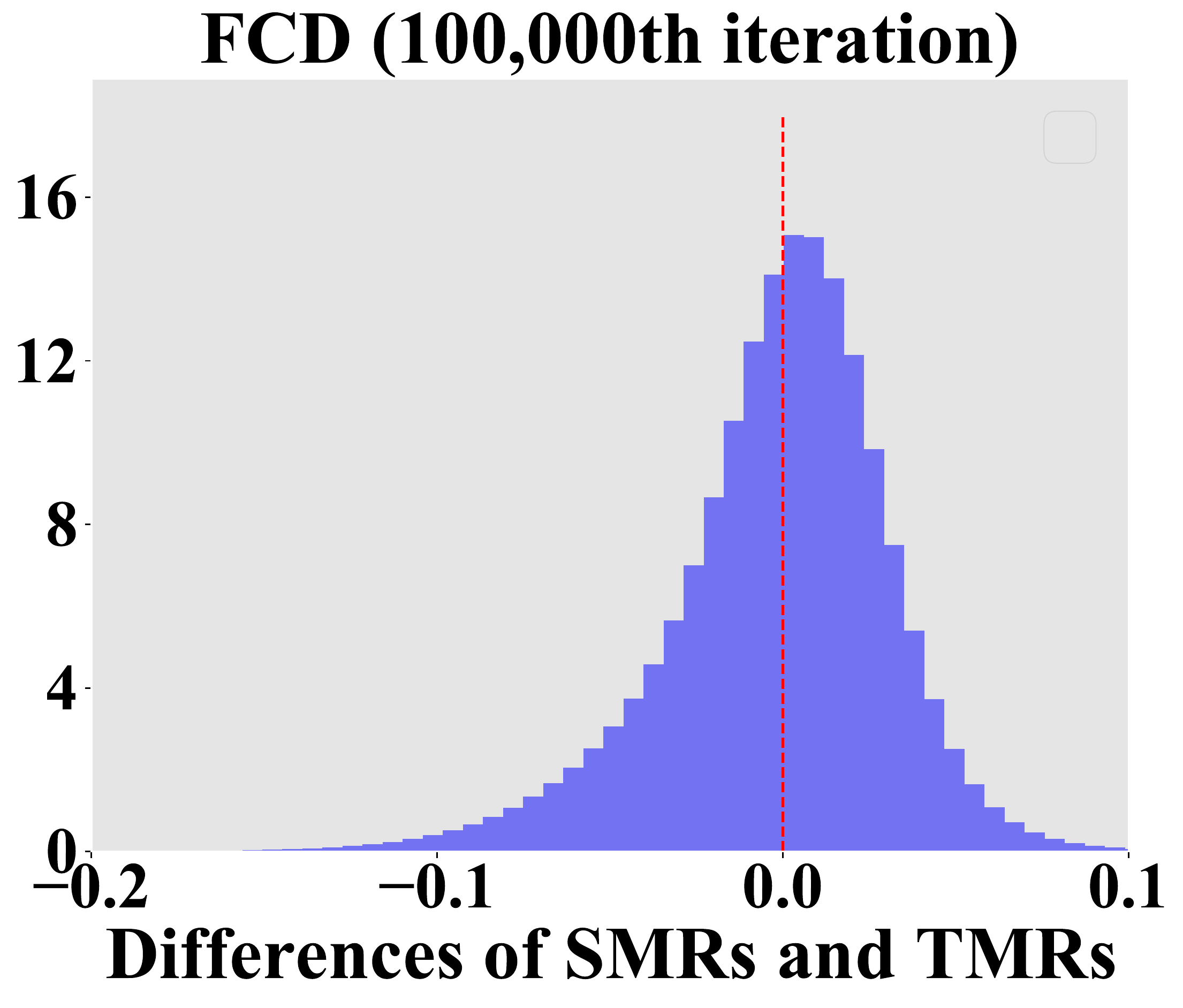}
        \label{label_for_cross_ref_3}
    }

    \subfigure{
        \includegraphics[width=0.25\linewidth]{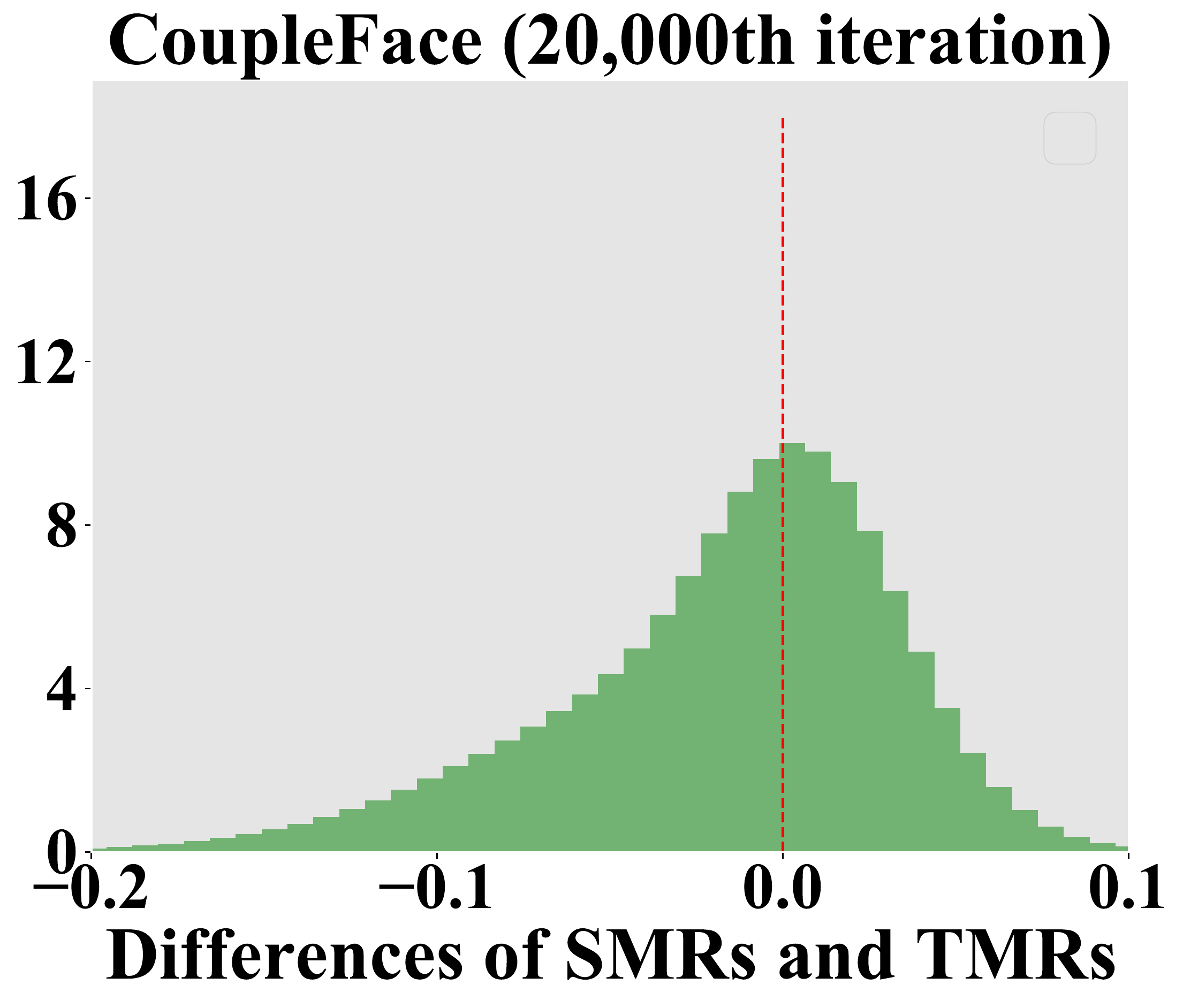}
        \label{label_for_cross_ref_4}
    }
    \subfigure{
	\includegraphics[width=0.25\linewidth]{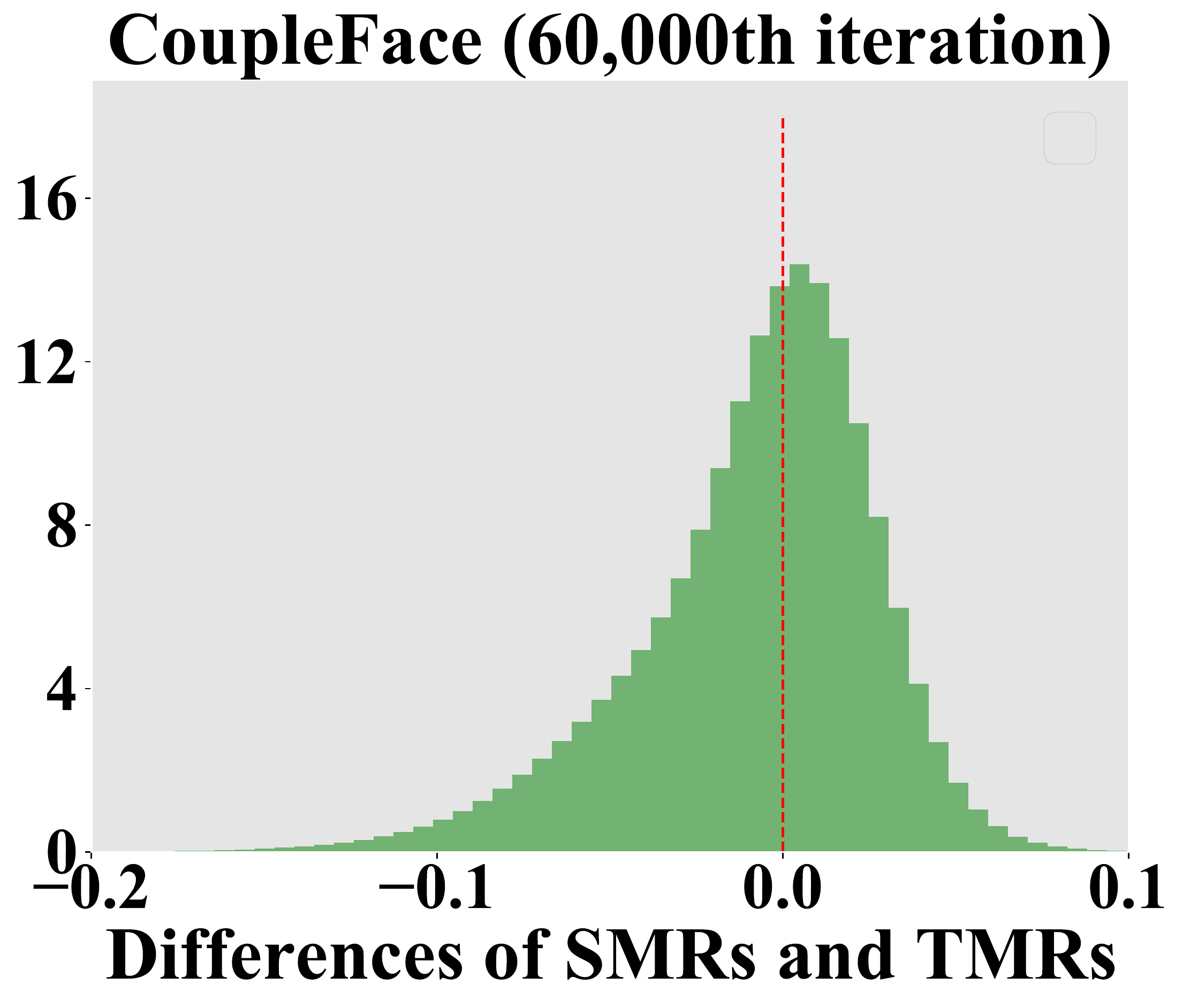}
        \label{label_for_cross_ref_5}
    }
    \subfigure{
    	\includegraphics[width=0.25\linewidth]{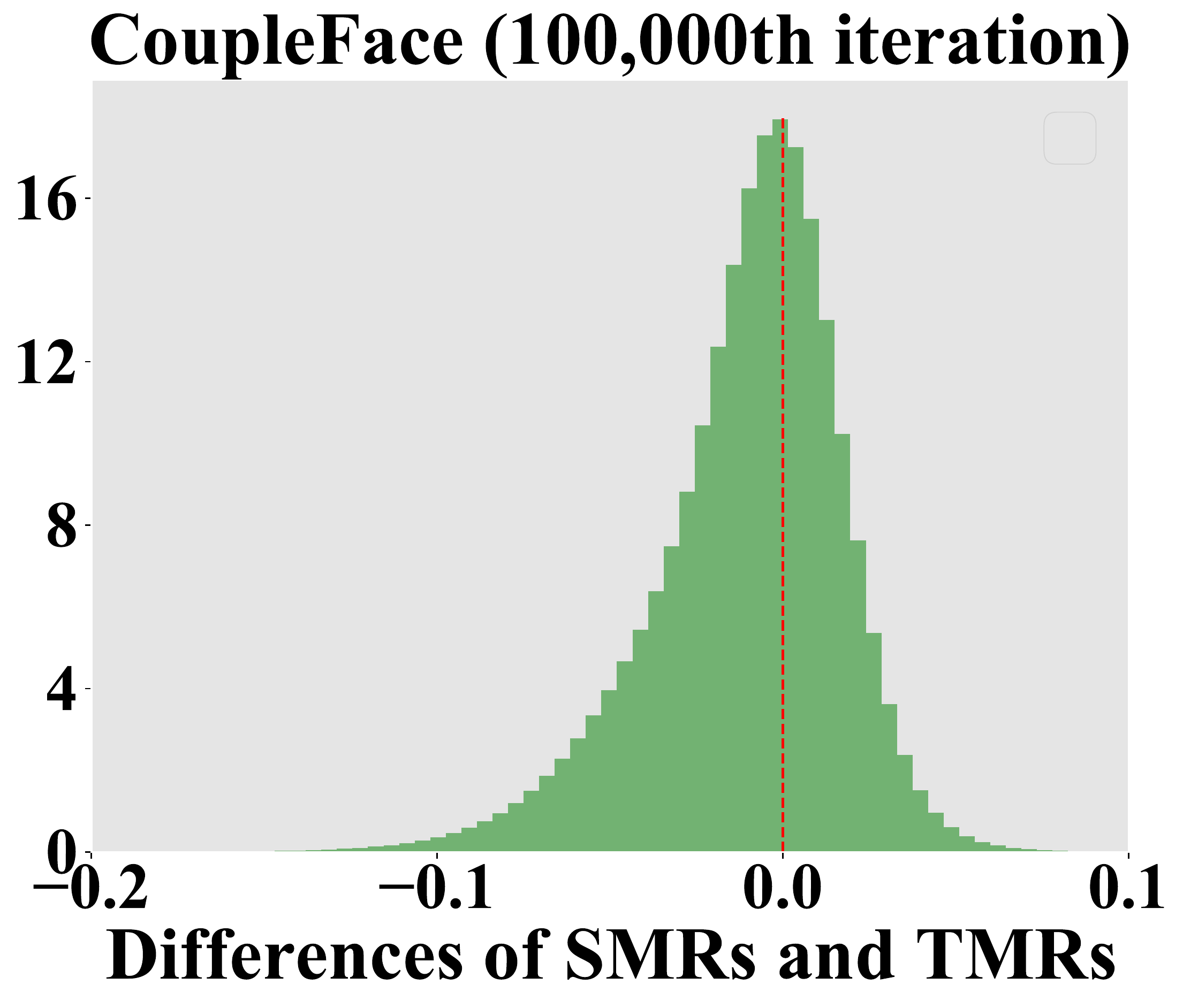}
        \label{label_for_cross_ref_6}
    }
    \caption{The distributions of differences between the SMRs and TMRs of different iterations for FCD and CoupleFace.}
    \label{vis}
\end{figure}

\noindent \textbf{Visualization on the differences of SMRs and TMRs.}
To further analyze the effect of CoupleFace,
we visualize the distributions of the differences between  SMRs and TMRs for both FCD and  CoupleFace in Fig.~\ref{vis}.
Specifically, we use the models of MBNet in the 20,000th, 60,000th, 100,000th iterations for both FCD and CoupleFace.
The first and the second rows show the results of FCD and CoupleFace,
respectively.
In Fig.~\ref{vis},
during training,
 for CoupleFace,
we observe that the differences of SMRs and TMRs gradually decrease at the right side of the red line,
which demonstrates the effect of minimizing the RAD loss of Eq.~(\ref{L_R3}).
Besides,
when compared with FCD,
most SMRs are less than or approximate to TMRs in CoupleFace,
which indicates that CoupleFace 
transfers
the mutual relation knowledge of the teacher model to the student model well.

\begin{figure}[t]
	\centering
	\subfigure{
		\includegraphics[width=0.35\textwidth]{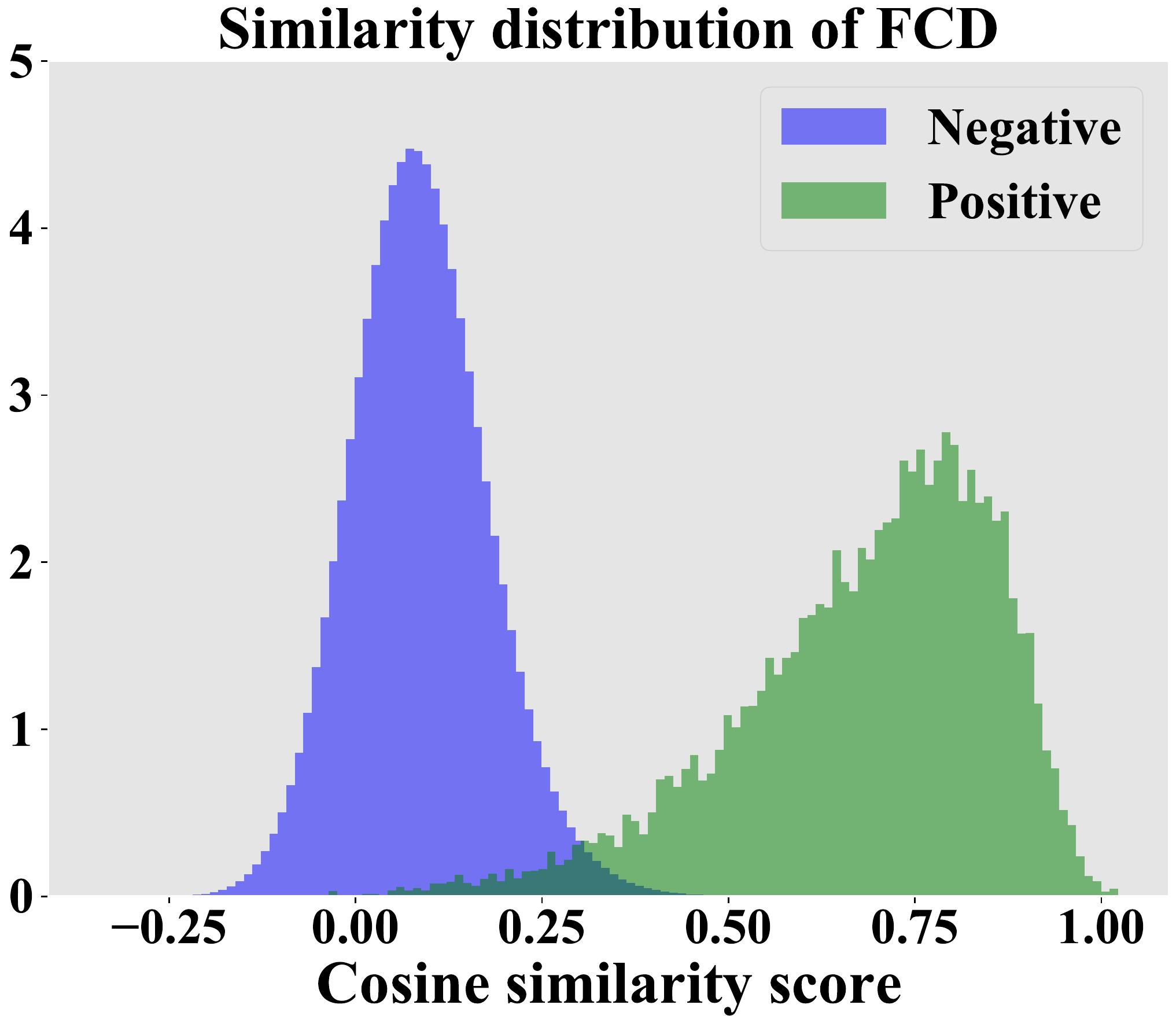}
	}
	\subfigure{
		\includegraphics[width=0.35\textwidth]{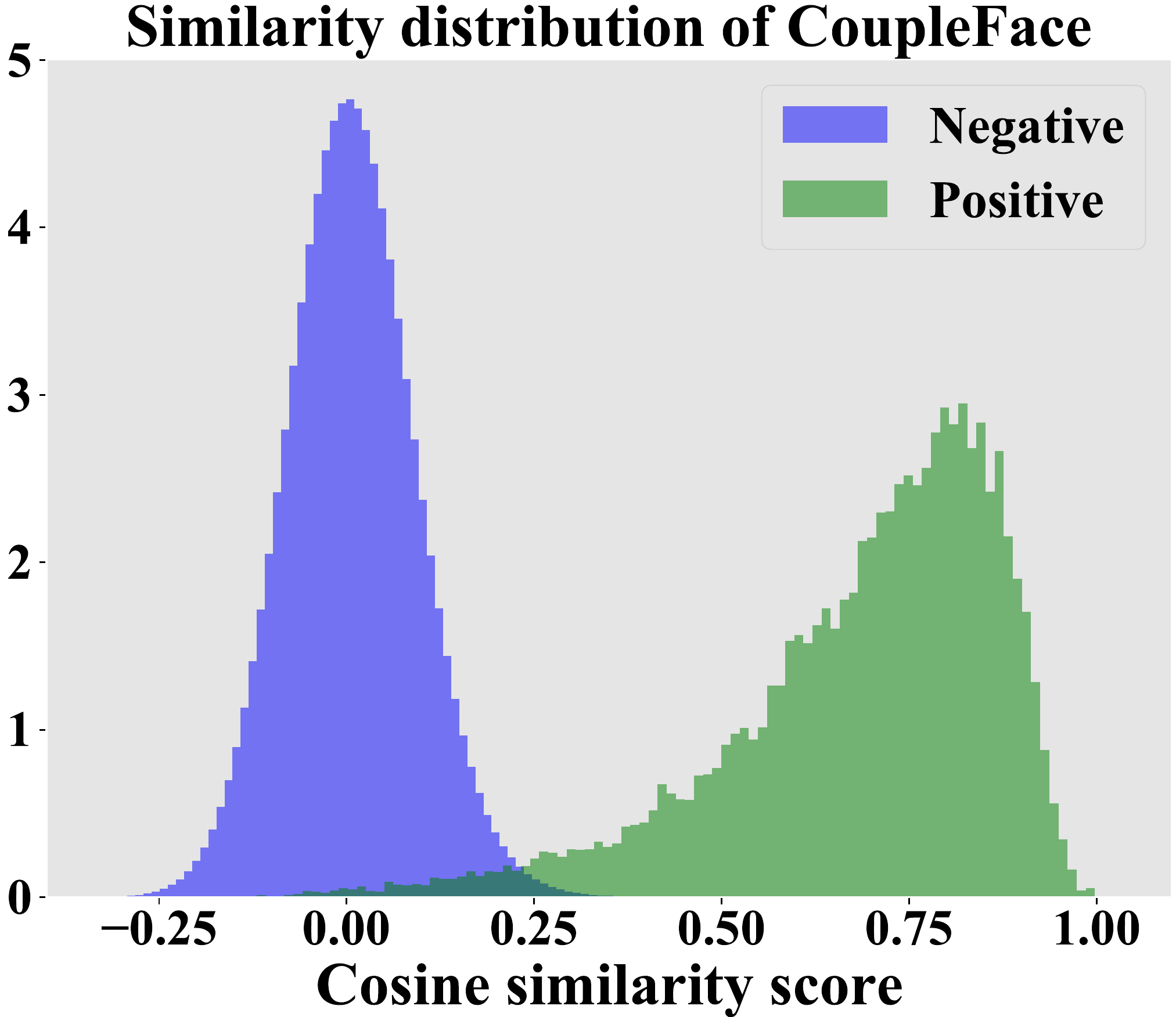}
	}
	\caption{Cosine similarity distributions of the positive pairs and negative pairs.}
	\label{fig:scores}
\end{figure}

\noindent \textbf{Visualization on the distributions of similarity scores.}
We visualize the distributions of similarity scores  on IJB-C of \textbf{MBNet} based on different methods in Fig.~\ref{fig:scores}.
Specifically,
we still use the \textbf{R-50} as $\mathcal{T}$ to distill \textbf{MBNet} in FCD and CoupleFace.
As shown in Fig.~\ref{fig:scores},
when compared with FCD, the similarity distributions of positive pairs and  negative pairs in CoupleFace are more compact and separable, which further shows the effectiveness of  CoupleFace.

\noindent\textbf{Comparison with existing relation-based knowledge distillation methods (RB-KDs)}. The differences between CoupleFace and existing RB-KDs (e.g., CCKD, RKD~\cite{Peng_2019_ICCV,park2019relational}) are as follows. (1) General KDs are usually incompatible with FR.
Existing RB-KDs are proposed for general vision tasks (e.g., close-set classification). In contrast, it is non-trivial to transfer relation knowledge for open-set FR well and CoupleFace is well-designed for FR distillation.
(2) Mining is considered.
We observe that most mutual relations cannot provide valuable knowledge and affect the similarity distribution for FR,
so how to produce sufficient informative mutual relations efficiently is a challenging issue.
In  CoupleFace,
we propose to mine informative mutual relations in MRD, while existing RB-KDs have not discussed the mining process.
(3) Loss function is intrinsically
 different. The RAD loss  in Eq.~(\ref{L_R3}) aims to better exploit the mutual relation knowledge by using valid mutual relations and filtering out the mutual relations with subtle differences, which are not discussed in existing works. (4) Better performance. CoupleFace outperforms existing RB-KDs a lot.
 
 \noindent \textbf{Computation costs.}
When compared with FCD, training time and GPU memory use of CoupleFace are {1.056}  times and {1.002} times, respectively.
which further demonstrates the efficiency of our proposed CoupleFace.
\section{Conclusion}
In our work, we  investigate the importance of mutual relation knowledge for FR distillation and propose an effective FR distillation method named as CoupleFace.
When compared with existing methods using  Feature Consistency Distillation (FCD),
CoupleFace further introduces the Mutual Relation Distillation (MRD),
where we  propose to mine the informative mutual relations and  utilize the Relation-Aware Distillation (RAD) loss to transfer the mutual relation knowledge from the  teacher model to the student model.
Extensive experiments on multiple FR benchmark datasets demonstrate the effectiveness of CoupleFace.
In our future work,
we will continue to  explore what kind of information is important for FR distillation and develop more effective distillation methods.

\section{Acknowledgments}
This research was supported by National Natural Science Foundation of China under Grant 61932002.

\clearpage
%
%
\bibliographystyle{splncs04}
\bibliography{egbib}
\end{document}